\documentclass[authoryear,preprint,review,12pt]{elsarticle}

\usepackage{amssymb}
\usepackage{amsfonts,amsmath,amscd}
\usepackage{float}
\usepackage{graphicx}
\usepackage{subfig}
\usepackage{nccmath}
\usepackage{siunitx}
\usepackage{booktabs}
\usepackage{url}
\usepackage{caption}

\journal{Transportation Research Part C: Emerging Technologies}

\begin{document}

\begin{frontmatter}



\title{Learning-Based Modeling of Human-Autonomous Vehicle Interaction for Improved Safety in Mixed-Vehicle Platooning Control}


\author[label1]{Jie\ Wang\corref{cor1}}
\ead{jwangjie@outlook.com}
\cortext[cor1]{Corresponding author}
\author[label1]{Yash Vardhan\ Pant}
\ead{yash.pant@uwaterloo.ca}
\author[label2]{Zhihao\ Jiang}
\ead{jiangzhh@shanghaitech.edu.cn}

\affiliation[label1]{organization={Electrical and Computer Engineering Department, University of Waterloo},%Department and Organization
            % addressline={200 University Avenue West}, 
            city={Waterloo},
            % postcode={N2L~3G1}, 
            state={ON},
            country={Canada}}

\affiliation[label2]{organization={School of Information Science and Technologies, ShanghaiTech University},%Department and Organization
            % addressline={393 Middle Huaxia Road}, 
            % city={Pudong},
            % postcode={201210}, 
            state={Shanghai},
            country={China}}

\begin{abstract}
The rising presence of autonomous vehicles (AVs) on public roads necessitates the development of advanced control strategies that account for the unpredictable nature of human-driven vehicles (HVs). This study introduces a learning-based method for modeling HV behavior, combining a traditional first-principles approach with a Gaussian process (GP) learning component. This hybrid model enhances the accuracy of velocity predictions and provides measurable uncertainty estimates. We leverage this model to develop a GP-based model predictive control (GP-MPC) strategy to improve safety in mixed vehicle platoons by integrating uncertainty assessments into distance constraints. Comparative simulations between our GP-MPC approach and a conventional model predictive control (MPC) strategy reveal that the GP-MPC ensures safer distancing and more efficient travel within the mixed platoon. By incorporating sparse GP modeling for HVs and a dynamic GP prediction in MPC, we significantly reduce the computation time of GP-MPC, making it only marginally longer than standard MPC and approximately 100 times faster than previous models not employing these techniques. Our findings underscore the effectiveness of learning-based HV modeling in enhancing safety and efficiency in mixed-traffic environments involving AV and HV interactions.
\end{abstract}

% %Graphical abstract
% \includegraphics[width=0.90\columnwidth]{figures/car_simulator.jpg}

% \begin{highlights}
% \item A Gaussian process (GP) based model corrects predictions and estimates uncertainty.
% \item An uncertainty-aware controller uses GP models to enhance vehicle interaction safety.
% \item The controller is 100x faster than previous methods, ensuring real-time control.
% \end{highlights}


\end{frontmatter}


\section{Introduction}
\label{section:Introduction}

\begin{figure}
    \centerline{\includegraphics[width=0.96\columnwidth]{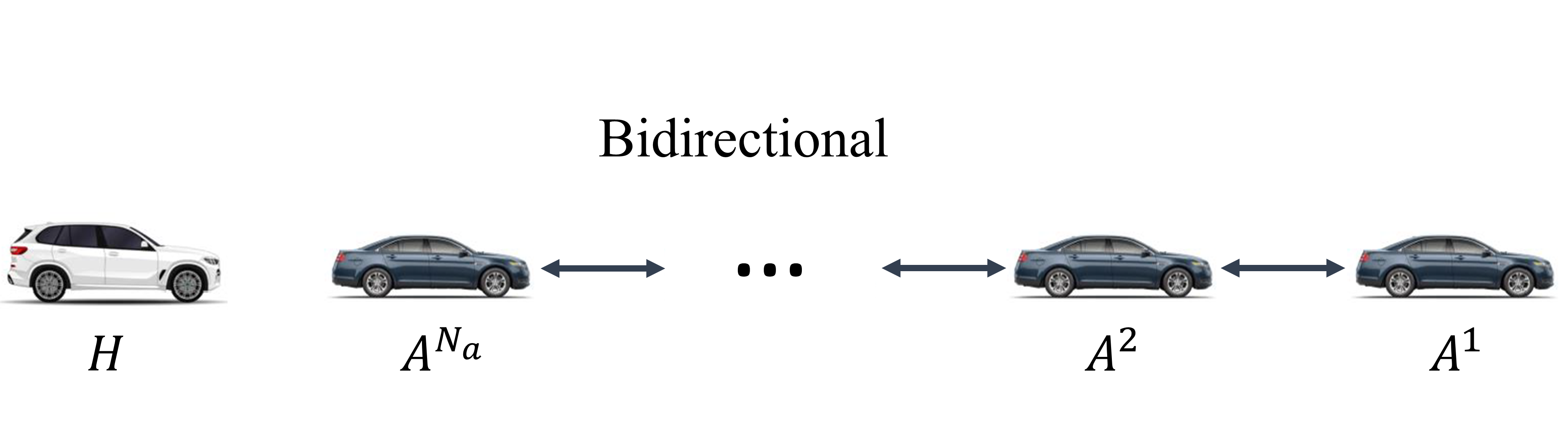}}
    \caption{Illustration of a mixed vehicle platoon comprising $N_a$ connected AVs, represented as ${A^1, A^2, \cdots, A^{N_a}}$, followed by a HV $H$. The AVs communicate in real-time using a sequential bidirectional communication topology, with no direct communication between CAVs and HVs. This differs from scenarios where an HV interacts with a single AV without front traffic cooperation, as AVs in a platoon can coordinate their actions to facilitate safer interactions between the following HV and the rear-most AV in the platoon. This configuration is motivated by recent studies highlighting that most accidents in mixed traffic involve HVs rear-ending AVs, leading to the specific platoon arrangement showcased.
    }
    \label{figure:1}
\end{figure}

The last decade has seen significant advancements in autonomous vehicles (AVs) and intelligent transportation systems, such as connected autonomous vehicle (CAV) platooning. These advancements aim to enhance traffic safety and efficiency \citep{guanetti2018}. AV platooning relies on synchronized driving, which allows for reduced spacing between vehicles while maintaining safe speeds. In a platoon, CAVs can share real-time data about speed and distance, enabling coordinated control and maneuvers. This synchronization allows for safe driving at higher speeds with smaller gaps. However, as human-driven vehicles (HVs) are expected to remain predominant on roads for the foreseeable future \citep{rahmati2019}, interactions between autonomous and human drivers are inevitable. An example of this is shown in Fig. \ref{figure:1}, where an HV is following a CAV platoon.

A recent study on traffic incidents in the US involving AVs found that 64.2\% of accidents in mixed traffic involved HVs rear-ending AVs, compared to only 28.3\% in HV-only traffic \citep{guo2020}. This increase in rear-end collisions is attributed to drivers' unfamiliarity with the driving dynamics of AV platoons. Instead of relying solely on human drivers to adapt to AVs, it is more practical and effective to develop CAV control strategies that account for human driving behaviors, ensuring safer cooperation in mixed-traffic environments. Although previous studies have examined HV--AV interactions for car-following scenarios, many existing models are not directly adaptable for model-based control applications \citep{sadat2020}. Therefore, there's a need for an HV--AV interaction model that is both comprehensible and directly applicable for designing model-based control strategies, with a focus on safety and other desired outcomes.

\subsection{Contributions}
This paper introduces a learning-based approach for modeling interactions between HVs and AV platoons in longitudinal car-following control scenarios. Our contributions are:

\begin{itemize}
    \item A novel method that combines a first-principles model of human behaviors and a learning-based Gaussian process (GP) model trained on human-in-the-loop simulator data to model HVs in a longitudinal velocity-tracking setting. The GP model corrects and estimates the uncertainty of the velocity prediction. The proposed model reduces the modeling error by 35.6\% compared to the first-principles model. To enhance the GP model's efficiency (full GP), we employed a sparse GP technique to reduce the average computational time by 18 times compared to the full GP, while still achieving a 24\% increase in modeling accuracy over the first-principles model.

    \item Development of a chance-constrained model predictive control (MPC) strategy, GP-MPC that maintains a safe distance in HV--AV interactions. The GP-MPC incorporates the proposed HV model to estimate modeling uncertainties as an additional probabilistic constraint. By combining this with a predefined deterministic distance, we establish an adaptive safe distance constraint that enhances safety.

    \item Demonstration of effectiveness of the GP-MPC over a traditional baseline MPC method with simulation experiments. Details of the baseline MPC are available in Sec. \ref{sec:simulations}. GP-MPC not only ensures a greater safety margin between vehicles but also enhances the efficiency of the mixed platoon by allowing higher vehicle speeds. Compared to the baseline MPC, the GP-MPC requires only a 5\% increase in computation time due to the use of sparse GP modeling for HVs and dynamic GP prediction in the MPC. As a result, the average computation time of the GP-MPC is approximately 100 times faster than previous work \citep{wang2024improving} that did not employ these techniques.   
\end{itemize}

\subsection{Related work}

In recent years, mixed traffic scenarios featuring both autonomous vehicles (AVs) and human-driven vehicles (HVs) have received increasing attention \citep{guo2020}. However, existing studies often presume uniform human behaviors or face challenges in incorporating HV uncertainties into model-based control policies. Traditional fixed-form parametric models often assume a fixed reaction delay for human drivers and generalize driving behaviors based on observations of similar drivers \citep{pirani2022}. While these models capture some aspects of human driving, their simplicity limits their accuracy.

Addressing these gaps, researchers have explored learning-based methods like Artificial Neural Networks (ANNs) such as multilayer neural networks, recurrent neural networks, and radial basis function networks\citep{morton2016}, as well as Hidden Markov Models \citep{qu2017}, and Gaussian Mixture Regression \citep{lefevre2014}. These methods offer enhanced accuracy in predicting human driving behavior but generally do not provide interpretable variance in modeling uncertainties. This limitation poses challenges for integrating these uncertainties into control policies and for performance and safety analyses. Conversely, Gaussian Process (GP) models, used in complex robotic systems for learning-based control, are advantageous for their fewer required parameters and their ability to offer interpretable uncertainties, crucial for safety analyses in control systems \citep{hewing2019}.

Within autonomous vehicle control, the GP-based MPC framework has become increasingly important. Research, such as that by \citep{hewing2019} and \citep{wang2023learning}, has applied GP models to refine AV modeling with historical data, improving MPC performance. In understanding AV interactions, studies like \citep{brudigam2021} and \citep{zhu2022} have used GP models to analyze behaviors of lead vehicles, especially for planning maneuvers in autonomous racing. These GP models improve overtaking strategies by predicting the lead vehicle's behavior and identifying overtaking opportunities based on perceived weaknesses. Furthermore, \citep{yoon2021} demonstrated the efficacy of a GP learning-based MPC framework in predicting cut-in vehicle trajectories by considering interactions with surrounding traffic. The study effectively handles lane-changing behaviors using real-world data, leading to a smoother and safer driving experience. 

Recent applications of GP models have extended to human-robot interactions. For instance, \citep{haninger2022} used GPs to understand human force during collaborative assembly tasks with robotic manipulators, designing an MPC to adapt to inferred task modes and objectives. While this approach facilitated human-robot collaboration and efficiency, it did not capitalize on quantified human force uncertainties for enhanced safety. Addressing the limitations in current HV--AV modeling, our study proposes a hybrid model combining a fixed-form model and a GP-based learning component. This model captures both deterministic and stochastic aspects of human behavior as observed in \citep{chen2010}. Integrating a GP component allows us to utilize uncertainties within the HV model, improving the safety of our MPC design for mixed-vehicle platooning control \citep{wang2023learning}.

Our prior work \citep{wang2024improving} demonstrated the effectiveness of this approach in HV modeling accuracy and mixed platoon safety. However, the high computational time (19 seconds per time step) limited its practicality for AV platoon control. In our current study, we address this by implementing a sparse GP method \citep{snelson2005} and integrating dynamic sparse GP prediction within the MPC \citep{hewing2019}, as detailed in sections \ref{section:sparse_gpr} and \ref{sec:dynamic_sparseGP_mpc}. These advancements significantly reduce computational complexity while achieving considerable improvements in overall system control performance, as shown in Sec. \ref{sec:simulations}.

The paper is organized as follows. Sec. \ref{sec:HV_modeling} details our novel HV modeling method. Sec. \ref{sec:controller} develops our MPC policy using the proposed HV model. Sec. \ref{sec:simulations} presents simulations comparing our MPC with a baseline MPC strategy. Finally, Sec. \ref{sec:conclusion} concludes the paper.

\section{Learning-based Modeling of Human-driven Vehicle}
\label{sec:HV_modeling}

\subsection{First-principles model of human-driven vehicles}
The delay in reaction time is a significant human factor that impacts the performance of operators \citep{pirani2022}. To account for the characteristics of human central nervous latencies, neuromuscular system, and other human and environmental factors, a transfer function with time delay has been formulated in \citep{macadam2003} to model the human response as:
\begin{equation} 
    G_{H}(s) \approx K \frac{1+T_{z} s}{1+2 \gamma T_{w} s+T_{w}^{2} s^{2}} e^{-T_{d} s}=\frac{\dot{P}_{H}(s)}{\dot{P}_{N}(s)}  \, , \tag{1} \label{eqn:TF_func}
\end{equation}
Here, $G_{H}(s)$ is the transfer function in the s-domain. The factor $K$ denotes the system's gain, indicating the proportional strength of the system's response. The term $T_{z}$ represents the zero time constant, $\gamma$ signifies the damping coefficient, and the variable $T_{w}$ indicates the system's natural frequency. The exponential term $e^{-T_{d} s}$ introduces a time delay in the system, with $T_{d}$ specifying the delay in the human driver's response. Lastly, $\dot{P}_{H}(s)$ and $\dot{P}_{N}(s)$ represent the Laplace transformations of the velocities for the HV $v_{k}^{H}$ and the front AV $v_{k}^{N_a}$, respectively.

The parameters of $G_{H}(s)$ can be uncovered through the gathered data. Furthering our mathematical exploration, we apply a second-order Padé approximation to the time delay component nestled within the transfer function \eqref{eqn:TF_func}. This application allows us to derive a discrete-time difference equation for the Auto-Regressive with Exogenous input (ARX) model. The equation is obtained by discretizing the transfer function using the methodologies described in \citep{wang2024improving}. The resulting equation is expressed as:
\begin{ceqn} 
    \begin{align}
        v^{H}_{k} &= -c_1 v^{H}_{k-1} - c_2 v^{H}_{k-2} - c_3 v^{H}_{k-3} - c_4 v^{H}_{k-4} \nonumber \\
        & \, \quad + b_1 v^{N_a}_{k-1} + b_2 v^{N_a}_{k-2} + b_3 v^{N_a}_{k-3} + b_4 v^{N_a}_{k-4} \, , \nonumber \\
        & = {f}\left(v_{k-1:k-4}^{H}, {v}_{k-1:k-4}^{N_a} \right) \, . \tag{2} \label{eqn:arx}
    \end{align}
\end{ceqn}
Here, $v^{H}_{k-i}$ and $v^{N_a}_{k-i}$ represent the velocities of the HV and the last vehicle in the AV platoon at time step $k-i$, respectively. In equation \eqref{eqn:arx}, the coefficients $c_1, c_2, c_3,$ and $c_4$ are the auto-regressive coefficients associated with the HV's velocities at previous time steps. Similarly, $b_1, b_2, b_3,$ and $b_4$ are the exogenous input coefficients tied to the velocities of the last vehicle in the AV platoon at preceding time intervals. These coefficients are determined by the discretization of $G_{H}(s)$ and reflect the impact of past velocities on the current velocity of the HV. The variable $k$ is the discrete time step or instance at which the velocities are being considered, while $i$ is an index used to refer to previous time steps, taking values from 1 to 4 in this context.

\subsection{The proposed modeling method for HVs}
Although HVs can be modeled as an ARX model in \eqref{eqn:arx} with limited accuracy by assuming a fixed reaction delay \citep{pirani2022}, incorporating GP models can improve prediction accuracy and provide uncertainty estimations that can be used as constraints for safety guarantees in control. Therefore, we propose modeling HVs in an ARX+GP format as:
\begin{subequations}
    \label{eq:arx_gp_model}
    \begin{align}
        v^H_k &= \sum_{i=1}^4 -c_i v^H_{k-i} + \sum_{i=1}^4 b_i v^{N_a}_{k-i}  = {f}(\cdot) \quad \text{ (rewrite \ref{eqn:arx})} \, , \tag{3a} \label{eqn:HV_nominal}\\
        \tilde{v}^{H}_{k}&=\overbrace{v^H_k}^{\text {ARX prediction}}+\overbrace{{g}(v^{H}_{k-1}, v^{N_a}_{k-1})} ^{\text {GP-based correction}} \, . \tag{3b} \label{eqn:HV_corrected_modeling}
    \end{align}
\end{subequations}
Here, $\tilde{v}^{H}_{k}$ denotes the GP-compensated velocity prediction of the HV. In the proposed framework, the model ${f}(\cdot)$ represents an ARX nominal model, characterized by constant parameters $b_i$ and $c_i$. Additionally, ${g}(\cdot)$ serves as a GP model, tasked with learning the divergences between the actual system behaviors (as gleaned from system data) and the nominal ARX model. Crucially, The GP model ${g}(\cdot)$ is dependent on both $v^{H}$ (HV velocity) and $v^{N_a}$ (the velocity of the last AV). This ensures that the GP model adequately captures the dynamics of the HV influenced by the actions of the leading AV.

\subsection{GP model training}
\label{section:sparse_gpr}

In equation \eqref{eqn:HV_corrected_modeling}, a GP model is employed to estimate the discrepancy between the actual velocities and the ARX-predicted velocities of HVs. Using previously gathered velocity state data points, namely $v^{H}_{j}$ and $v^{N_a}_{j}$, the discrepancies are learned by the GP model ${g}(\cdot)$ as expressed as:
\begin{equation}
    \hat{v}^{H}_{j} -{v}^{H}_{j} = {g}(\mathbf{a}_{j-1}) \, ,  \tag{4} \label{eqn:gp_data_prep}
\end{equation}
Here, the discrepancy state is denoted as $\mathbf{a}_{j-1}=(v^{H}_{j-1}, v^{N_a}_{j-1})$. Both $\hat{v}^{H}_{j}$ and ${v}^{H}_{j}$ refer to velocities ascertained from the collected data and predicted by the ARX model \eqref{eqn:HV_nominal}, respectively. Within this context, $\mathbf{a}_{j}$ stands for an individual data point within the observed input-output dataset.

In each experiment, all input-output data (one-dimensional) pairs are collected at every time step. These data points are then prepared according to equation \eqref{eqn:gp_data_prep} to create a discrepancy data set, represented as $\mathcal{D} = \{\mathbf{a} = [{a}_1, \cdots, {a}_n]^\mathsf{T} \in \mathbb{R}^{n_a \times n}, \mathbf{g} = [{g}_1, \cdots, {g}_n]^\mathsf{T} \in \mathbb{R}^{1 \times n}\}$, where $n$ stands for the total number of data in the experiment. During the GP model training/estimation phase, the kernel function hyperparameters are optimized by maximizing the log marginal likelihood of the collected disturbance observation data \citep{rasmussen2006}. As in various robotics control utilizing GP-based techniques \citep{hewing2019}, a squared exponential kernel was chosen for this work, as illustrated as: 
\begin{ceqn}
    \begin{equation} 
        k({a}_i, {a}_j) = \sigma_{f}^2 \exp \left[-\frac{1}{2}
        \left({a}_i - {a}_j \right)^\mathsf{T} \mathbf{L}^{-1}
        ({a}_i - {a}_j) \right] \, , \tag{5} \label{eqn:rbf}
    \end{equation}
\end{ceqn}
where ${a}_i$ and ${a}_j$ represent different data points within $\mathbf{a}$, each possessing a dimension of $n_a$-by-$1$. The hyperparameters $\sigma_{f}^2$ and $\mathbf{L} \in \mathbb{R}^{n_a \times n_a}$ are optimized during the training process of the GP models using the observed dataset \citep{wang2023intuitive}. Given the observed data $\mathcal{D}$, we obtain a predictive posterior distribution to approximate the unknown function by a trained GP model ${g}(\cdot)$. Consequently, predictions can be made at a new data point ${a}^*$, which has the same dimension as each observed data point ${a} \in \mathbb{R}^{n_a \times 1}$, as described in \citep{rasmussen2006} through 
\begin{ceqn} 
    \begin{equation} 
        {g}({a}^*) \sim \mathcal{N} \left({\mu}^{{d}}({a}^*), {\Sigma}^{{d}}({a}^*) \right) \, ,  \tag{6} \label{eqn:gp_eqn}
    \end{equation}
\end{ceqn}
where 
\begin{ceqn}
    \begin{align} 
       \mu^d({a}^*) &= \mathbf{K}_{{{a}^*}\mathbf{a}} \left(\mathbf{K}_{\mathbf{a}\mathbf{a}} + \sigma^2 \mathbf{I} \right)^{-1} \mathbf{g} \, , \tag{7a} \label{eqn:gp_mean}   \\ 
       \Sigma^d({a}^*) &= \mathbf{K}_{{{a}^*}{{a}^*}} -  \mathbf{K}_{{{a}^*}\mathbf{a}}\left(\mathbf{K}_{\mathbf{a}\mathbf{a}} + \sigma^2 \mathbf{I} \right)^{-1} \mathbf{K}_{\mathbf{a}{{a}^*}} \, . \tag{7b} \label{eqn:gp_var}
    \end{align}
\end{ceqn} 
In this case, the GP model has a zero mean and a prior kernel $k(\cdot, \cdot)$, and $\mathbf{K}_{\mathbf{a}\mathbf{a}}$ denotes the Gram matrix obtained by applying the kernel function $k(\cdot, \cdot)$ to the observed input data $\mathbf{a}$, i.e., $[\mathbf{K}_{\mathbf{a}\mathbf{a}}]_{ij} = k({a}_i, {a}_j)$. Moreover, $[\mathbf{K}_{\mathbf{a}{a}^*}] = k({a}_j, {a}^*)$, $\mathbf{K}_{{{a}^*}\mathbf{a}} = (\mathbf{K}_{\mathbf{a}{{a}^*}})^\mathsf{T}$, and $\mathbf{K}_{{{a}^*}{{a}^*}} = k({a}^*,{a}^*)$. 

The computational complexity of the mean \eqref{eqn:gp_mean} and variance \eqref{eqn:gp_var} are $\mathcal{O}(n_a m)$ and $\mathcal{O}(n_a m^2)$ respectively, and increase with the training data points number $m$ \citep{hewing2019}. To reduce the computational complexity of the standard GP model \eqref{eqn:gp_eqn} while maintaining reasonable approximation accuracy, we applied one state-of-the-art sparse spectrum method, the fully independent conditional (FIC) approximation \citep{snelson2005}. 
The underlying concept is to compress the information contained in the training data into a subset of the data as inducing variables, allowing for the storage of only these variables instead of the complete dataset. The FIC method automatically determines the locations of these inducing variables by maximizing the GP marginal likelihood by gradient ascent.

% \section{HUMAN-DRIVEN VEHICLE MODEL}
\subsection{ARX+GP model}
\label{sec:HV_model}
\begin{figure}
    \centering
    \vspace{0.1cm}
    \includegraphics[trim=0cm 0cm 0cm 0cm, width=0.94\columnwidth]{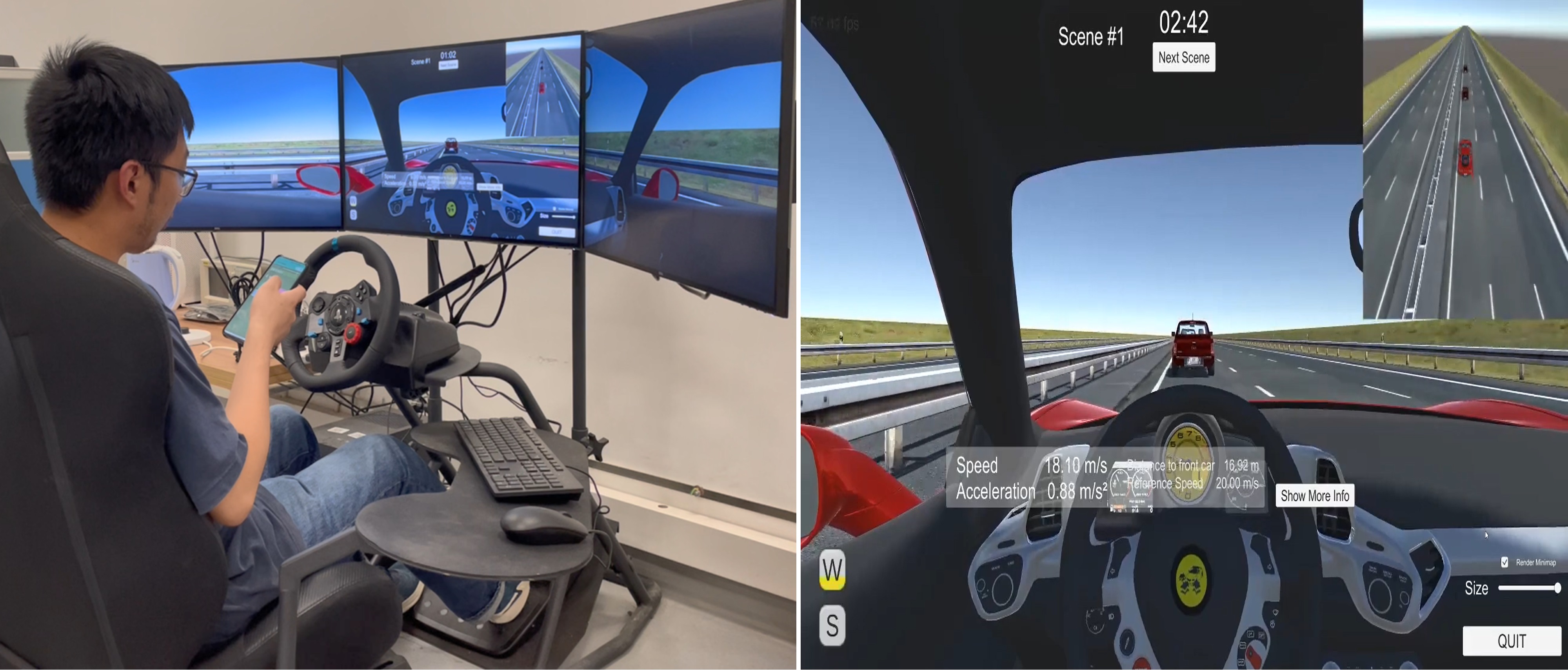}
    \caption{Data collection setup in a Unity driving simulator, featuring three drivers navigating with distractions across three different driving scenarios. In each trial, the human driver's objective was to tail a platoon of two AVs for a span of 3 minutes. The leading AV adhered to a reference velocity profile segmented into a constant 20\,m$/$s for 30\,s, a linear chirp waveform ranging from 25 to 10\,m$/$s over 60\,s, and a constant 10\,m$/$s for 90\,s.
    }
    \label{figure:car_simulator}
\end{figure}

To estimate the HV model, we collected data from three distinct driving scenarios in which three drivers followed a platoon of two AVs within a Unity simulation environment, as shown in Fig. \ref{figure:car_simulator}.
During these experiments, the drivers were purposely distracted with a cognitive task - solving algebra questions with multiple choices - while tailing the AV platoon \footnote{See \url{https://youtu.be/mfPzYDQrvV4} for video playback.}. Our focus was on investigating human behavior modeling in such distracted driving scenarios, as they pose a substantial challenge to maintaining safe control in AV--HV interactions due to increased uncertainty in the HV model. Each experiment had the driver follow the AV platoon for 3 minutes, with the lead AV following a reference velocity profile composed of three segments: a constant 20\,m$/$s for 30\,s, a linear chirp waveform ranging from 25 to 10\,m$/$s over 60\,s, and a constant 10\,m$/$s for 90\,s \citep[Sec. V.B]{pirani2022}. 
The collected data points were analyzed by calculating their mean values. These mean values were then utilized to derive the transfer function described by equation \eqref{eqn:TF_func}. By referencing equation \eqref{eqn:arx}, the parameters of the ARX nominal model can be computed. 

Next, we utilized Equation \eqref{eqn:gp_data_prep} to prepare the data for estimating the discrepancy between the identified ARX model and the actual behavior data of the HV. Given the considerations of computational efficiency and data diversity, we made the informed decision to train the GP model using only 20\% of the data points. These were uniformly selected from six datasets, while the remaining three datasets were earmarked for testing purposes. During our experimentation phase, it became evident that training the GP model with all collected datasets resulted in extended durations, spanning several days. Such extended training timelines are not just impractical but also counterproductive, particularly when aiming to optimize the sparse GP model's capabilities. Additionally, the computational overhead associated with the integration of the ARX+GP model in a human-in-loop platooning control environment is significant. A pivotal observation was the lack of diversity in our current dataset, with a predominant concentration of data points at velocities of 10 m$/$s, 15 m$/$s, and 20 m$/$s. As such, simply increasing the volume of data points for training did not yield a significant improvement in the GP model's performance.

Upon training the GP model, we conducted tests with the remaining three datasets. The ARX and ARX+GP model velocity predictions, along with twice the standard deviation (2$\sigma$) estimated by the GP model, were plotted in Figure \ref{figure:gp_standard}. We also included the actual measured HV velocities in the graph to demonstrate the improvement of accuracy achieved with the GP model.
\begin{figure}
    \centering
    \captionsetup{labelformat=empty}
    \vspace{0.15cm}
    \subfloat[GP model evaluation on test dataset from driving scenario 1 with driver 2]{{\includegraphics[trim=0cm 0cm 0cm 0cm, width=0.96\columnwidth]{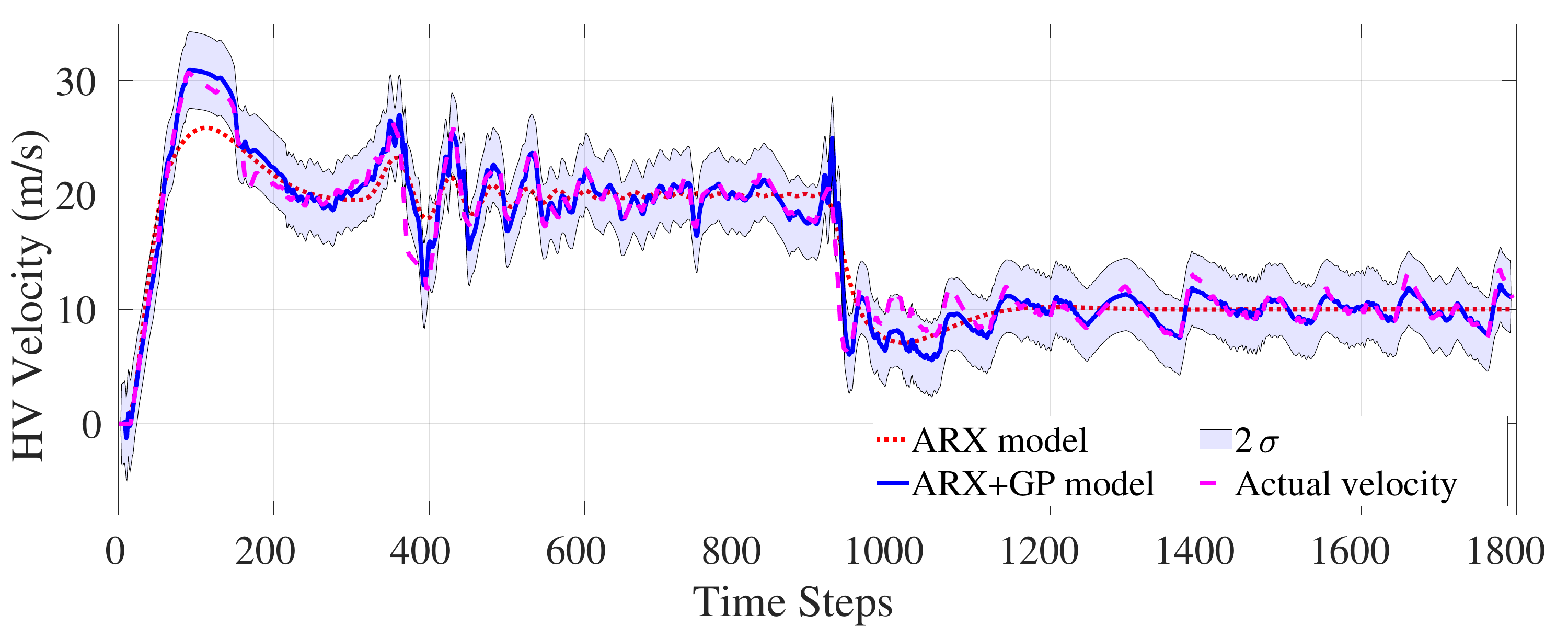} }}
    \qquad \qquad 
    \vspace{0.1cm}
    \subfloat[GP model evaluation on test dataset from driving scenario 2 with driver 1]{{\includegraphics[trim=0cm 0cm 0cm 0.0cm, width=0.96\columnwidth]{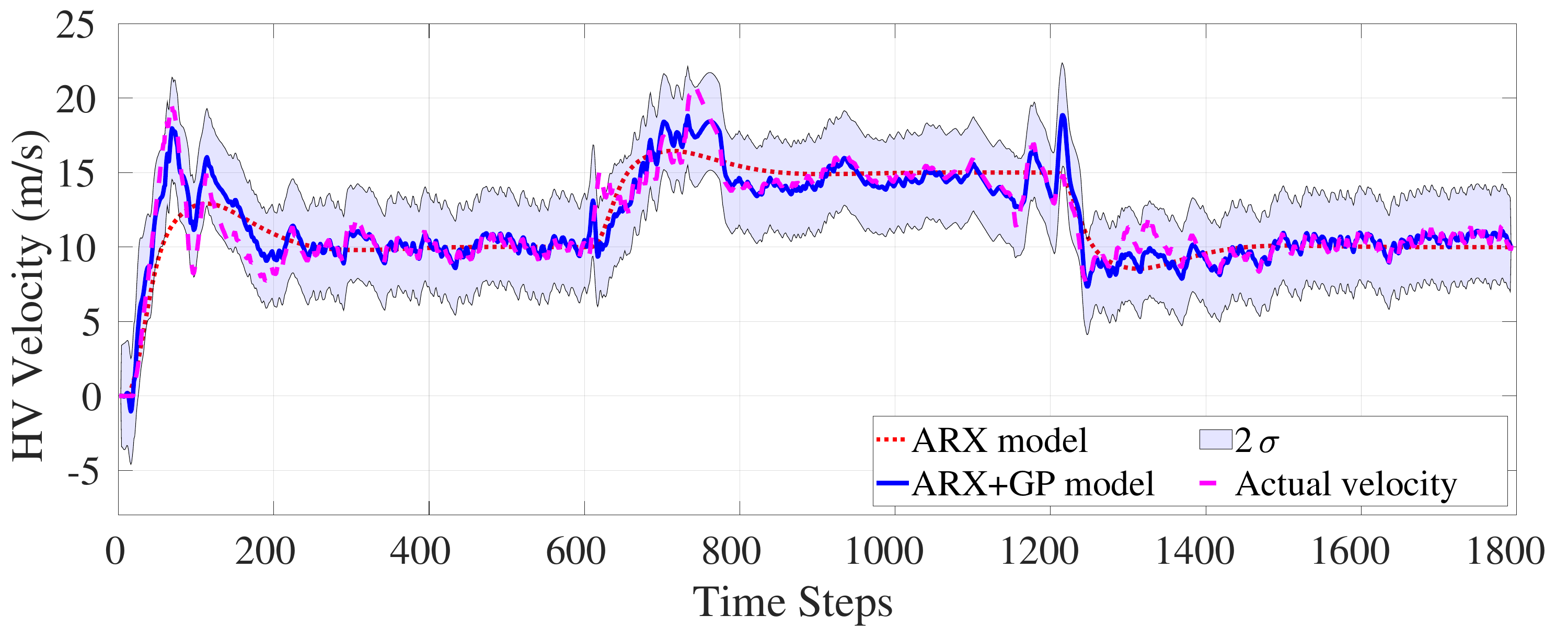} }}
    \qquad \qquad 
    \vspace{0.1cm}
    \subfloat[GP model evaluation on test dataset from driving scenario 3 with driver 1]{{\includegraphics[trim=0.0cm 0cm 0.0cm 0.0cm, width=0.96\columnwidth]{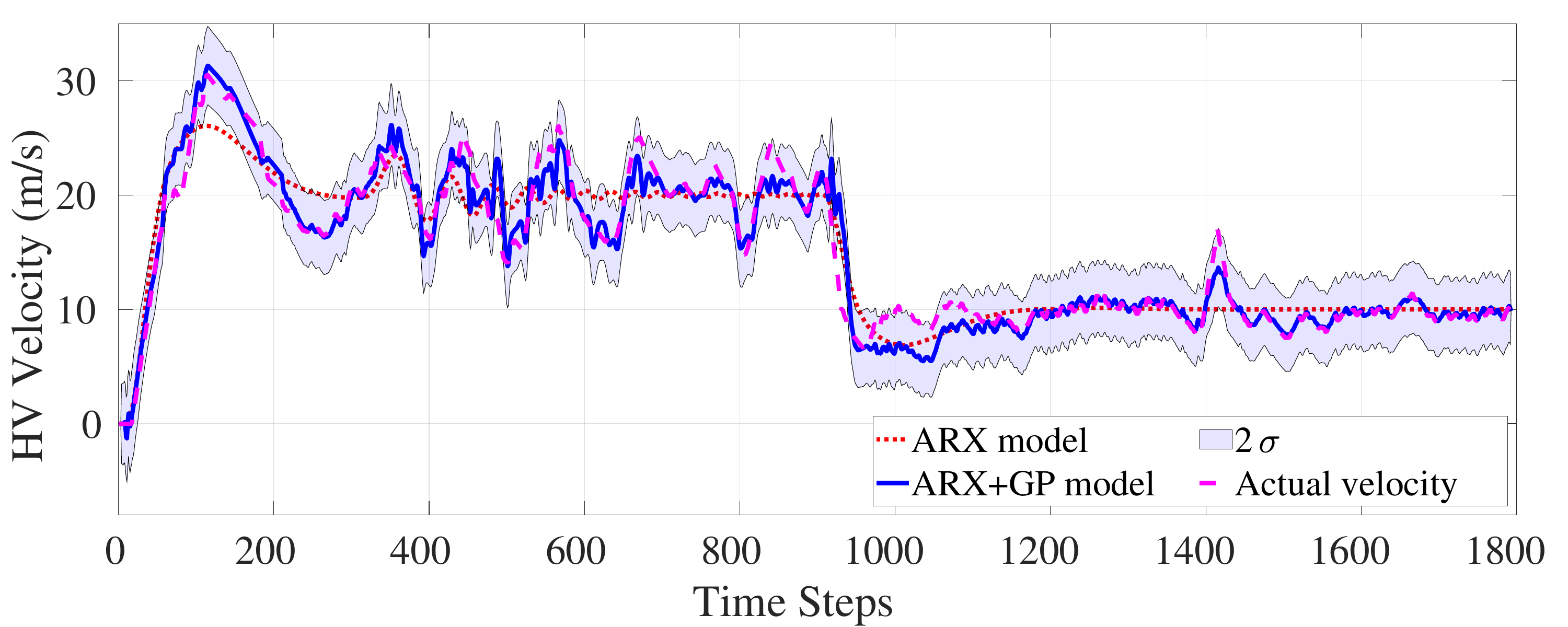} }}
    \caption{}
\end{figure}
\begin{figure}
    % \captionsetup{labelformat=adja-page}
    \ContinuedFloat
    \caption{The trained GP model's performance was evaluated using three testing datasets. The ARX and ARX+GP model's HV velocity predictions were compared by plotting them alongside the measured velocities and twice the standard deviation (2$\sigma$) determined using the GP model. The results highlight that the ARX+GP model substantially enhances the fit of velocity curves compared to the standalone ARX model. On average, the ARX+GP model improves modeling accuracy by 35.64\% in terms of RMSE compared to the ARX model.}
    \label{figure:gp_standard}
\end{figure}
To quantify the accuracy improvement, we calculated the root mean square error (RMSE) for each testing dataset using
\begin{ceqn}
    \begin{align}
        \text{RMSE}_{{v}^{H}_*}=\sqrt{\sum_{i=1}^n \frac{\left({v}^{H}_*-{v}^{H}_\text{act}\right)^2}{N}} \, , \tag{8} \label{eqn:rmse}
    \end{align}
\end{ceqn}
where ${v}^{H}_*$ represents either the ARX or the ARX+GP model's velocity predictions over $N$ samples in a testing dataset, and ${v}^{H}_\text{act}$ denotes the actual measured velocities. The RMSE measures the average magnitude of the prediction errors in the model across the testing datasets, so lower RMSE values indicate a higher level of accuracy in modeling the HVs. The ARX and ARX+GP model's RMSE results are summarized in Tab. \ref{tab:rmse_gp}. The average RMSE of the ARX model was found to be 1.88\,m$/$s, while the ARX+GP model achieved an average RMSE of 1.21\,m$/$s. This indicates an overall improvement in modeling accuracy of approximately 35.64\% for the ARX+GP model compared to the ARX model. The enhanced modeling accuracy is evident in Fig. \ref{figure:gp_standard}, where the ARX+GP model closely aligns with the actual velocity data curves compared to the ARX model. This improvement is particularly evident during time steps 200-600, 800-1200, and 1400-1800, where the velocities closely align with the training data points, especially around 10\,m$/$s or 15\,m$/$s.

\begin{table}
    \caption{RMSE results of the ARX, GP+ARX, and sparse GP+ARX models with three testing datasets.}
    \label{tab:rmse_gp}
    \centering
    \begin{tabular}{|c|c|c|c|}
        \hline
        \textbf{Testing Dataset} & \textbf{Dataset 1} & \textbf{Dataset 2} & \textbf{Dataset 3} \\
        \hline
        ARX model & 1.9025\,m$/$s & 1.5327\,m$/$s & 2.2054\,m$/$s \\
        \hline
        GP+ARX model & 1.2007\,m$/$s & 0.9815\,m$/$s & 1.4713\,m$/$s \\
        \hline
        Sparse GP+ARX model & 1.4917\,m$/$s & 1.2206\,m$/$s & 1.5950\,m$/$s \\
        \hline
    \end{tabular}
\end{table}

\subsection{Sparse GP+ARX model}
\label{sec:sparse_hv}
We have implemented the Fully Independent Conditional (FIC) sparse approximation technique to expedite the predictive capacity of our GP model. The hyperparameters for this sparse GP model were determined from the full GP models, as delineated in Section \ref{section:sparse_gpr}. The FIC method autonomously designated 20 inducing points within our training datasets. The performance of the sparse GP model was subsequently evaluated using three separate testing datasets. Remarkably, the GP model prediction time was decreased to just 0.00021\,s, a staggering 18 times quicker than the traditional full GP model, which required 0.0037\,s to finalize each prediction.

\begin{figure}
    \centering
    \vspace{0.15cm}
    \includegraphics[trim=0cm 0cm 0cm 0cm, width=0.98\columnwidth]{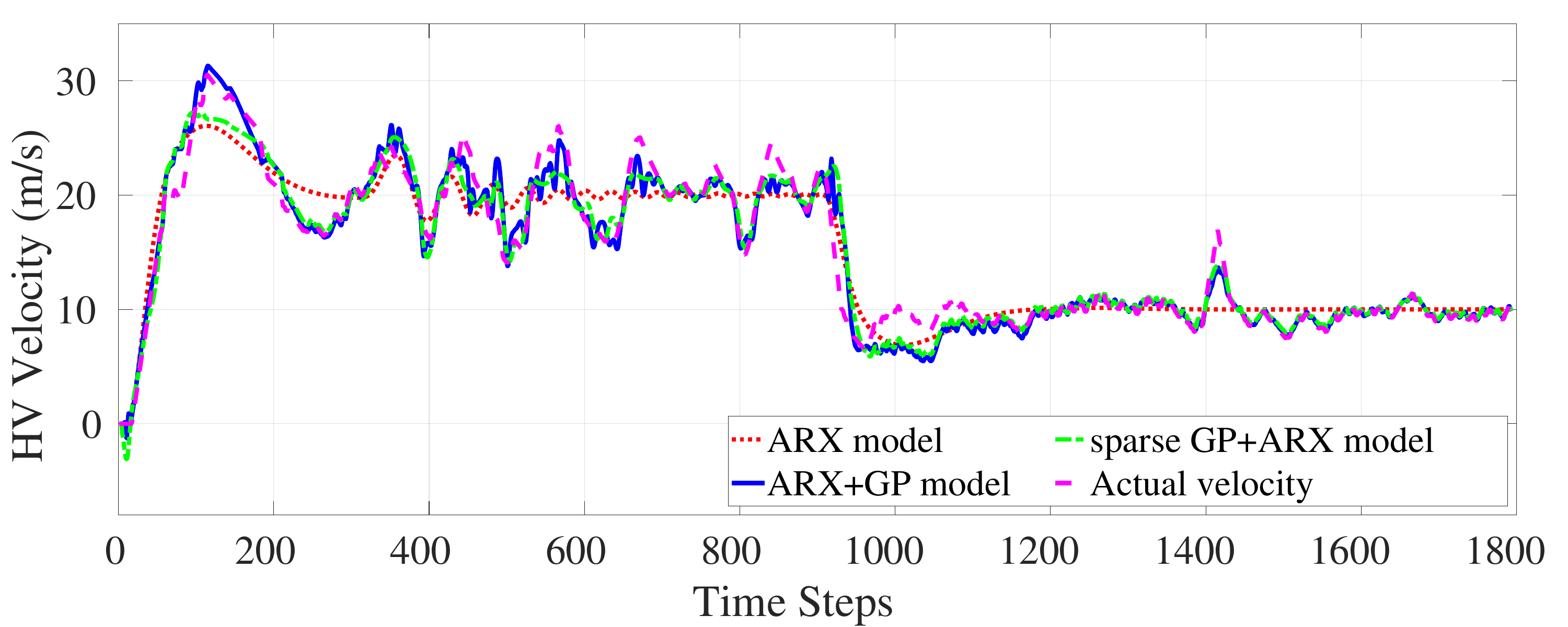}
    \caption{Performance results of the sparse GP model evaluated on one of three testing datasets. Plots show velocity predictions of the sparse GP+ARX, ARX+GP, and ARX models, alongside the measured HV velocities. Impressively, the sparse ARX+GP model continues to display a superior alignment with the genuine velocity curves, augmenting accuracy by an average of 23.94\%.
    }
    \label{figure:gp_sparse}
\end{figure}

For illustrative purposes, we incorporated a plot of one of the three test results from the sparse GP+ARX model (which corresponds to the bottom testing dataset depicted in Fig. \ref{figure:gp_standard}), displayed as Fig. \ref{figure:gp_sparse}. In this plot, we have plotted the velocity predictions of the ARX model, the ARX+GP model, and the sparse GP+ARX model together. We observed that the sparse GP+ARX model was also able to fit the actual velocity data curves much better than the ARX model. Moreover, by utilizing \eqref{eqn:rmse}, the RMSE values of the sparse GP+ARX model were summarized in Tab. \ref{tab:rmse_gp}. We also calculated the average RMSE to be 1.43\,m$/$s, indicating an approximate 23.94\% overall modeling accuracy improvement when compared to the ARX model. These results indicate that the sparse GP+ARX model provides a good balance between modeling accuracy and significant computational efficiency improvement.

\section{Controller Design}
\label{sec:controller}

Our proposed modeling approach's effectiveness for HVs is further demonstrated through a model predictive control strategy designed specifically to utilize this model. This demonstration highlights the effectiveness of the proposed approach in improving both performance and safety in the control of a mixed-vehicle platoon. The suggested method is particularly tailored for longitudinal car-following situations where a human-operated vehicle trails an autonomous vehicle platoon and does not apply to other interactions between human vehicles (HV) and autonomous vehicles (AV), such as lane-changing maneuvers.

Before delving into our proposed controller design, it's pertinent to contextualize our choices, especially when compared to the conventional adaptive cruise control (ACC) based on the constant time headway policy \citep{wu2020}. While ACC maintains a constant temporal gap to the lead vehicle and is efficient in stable traffic scenarios, it operates largely on deterministic inputs without inherently accounting for uncertainties. On the other hand, our GP learning-based MPC (GP-MPC) strategy, inspired by the research of \citep{hewing2019}, explicitly incorporates modeling uncertainties. It integrates both probabilistic constraints and a deterministic distance, offering enhanced safety assurances in uncertain driving scenarios. Though ACC is straightforward and widely adopted, the GP-MPC's adaptability makes it better suited for the complexities and uncertainties intrinsic to Human Vehicle (HV) and Autonomous Vehicle (AV) interactions, especially in the context of longitudinal car-following situations. 

\subsection{System model}

Visualize a mixed vehicle platoon consisting of AVs and a single HV, referred to as $A^{\mathbf{n_a}}$ and $H$, respectively. Here, $\mathbf{n_a} = \{1, 2, \cdots, N_a\}$, with $N_a$ signifying the quantity of AVs. At the present time step $k$, the velocity and position of $A^{\mathbf{n_a}}$ are symbolized by $v^{\mathbf{n_a}}_k$ and $p^{\mathbf{n_a}}_k$, respectively. The kinematic model of AVs is as follows:
\begin{ceqn}
    \begin{align}  
    v_{k+1}^{\mathbf{n_a}} &= v_{k}^{\mathbf{n_a}} + T \, \mathrm{acc}_{k}^{\mathbf{n_a}} \, , \tag{9a} \label{eqn:av_eqn_a} \\
    p_{k+1}^{\mathbf{n_a}} &= p_{k}^{\mathbf{n_a}} + T \, v_{k}^{\mathbf{n_a}} \, . \tag{9b} \label{eqn:av_eqn_b}
    \end{align}
\end{ceqn}
In the above, $0<T \ll 1$ indicates the sample time, and $\mathrm{acc}_{k}^{\mathbf{n_a}}$ signifies the acceleration of $A^{\mathbf{n_a}}$. We operate under the assumption that AVs are deterministic, measuring and communicating their states flawlessly, i.e., $\Sigma(v_{k}^{\mathbf{n_a}}) = 0$. Upon application of \eqref{eqn:HV_corrected_modeling}, the human-operated vehicle model becomes:
\begin{ceqn} 
    \begin{align} 
        \tilde{v}_{k}^{H} &= v_{k}^{H} + {g}(v_{k-1}^{H}, {v}_{k-1}^{N_a}) \, , \tag{10a} \label{eqn:sys_model_a} \\
        p_{k+1}^{H} &= p_{k}^{H} + T \, \tilde{v}_{k}^{H} \, , \tag{10b} \label{eqn:sys_model_b}
    \end{align}
\end{ceqn}
Here, $\tilde{v}_{k}^{H}$ represents the compensated HV velocity obtained from the GP model, while $v_{k}^{H}$ (computed using \eqref{eqn:arx}) and $p_{k}^{H}$ denote the velocity and position states at the current time step $k$. At the current time, the variances $\Sigma(v_{k}^{H})$ and $\Sigma(p_{k}^{H})$ are both equal to zero. By employing \eqref{eqn:gp_eqn}, we can propagate the mean of the position using the following equation:
\begin{ceqn} 
    \begin{align} 
        \mu_{k+1}^{p^{H}} &= \mu_{k}^{p^{H}} + T \, v_{k}^{H} + T \, \mu^d (v_{k-1}^{H}, {v}_{k-1}^{N_a}) \, . \tag{11} \label{eqn:mean_prop}
    \end{align}
\end{ceqn}
with the initial value is $\mu_{0}^{p^{H}} = p_{k}^{H}$. We also have the propagation equation for the position variance as:
\begin{ceqn} 
    \begin{align} 
        \Sigma \left(p_{k+1}^{H} \right) &= \Sigma \left( p_{k}^{H} + T \, v_{k}^{H} + T \, {g}(v_{k-1}^{H}, {v}_{k-1}^{N_a}) \right) \, , \tag{12a} \label{eqn:variance_prop_a} \\
        & = \Sigma_{k}^{p^{H}} + T^2 \Sigma^d_{k-1} \ , \tag{12b} \label{eqn:variance_prop_b}
    \end{align}
\end{ceqn}
The initial value is $\Sigma_{0}^{p^{H}} = 0$, and \eqref{eqn:variance_prop_b} overlooks the covariance between $p_{k}^{H}$ and $v_{k}^{H}$.

\subsection{Safe distance probability constraint}
To maintain a secure distance within the mixed platoon of vehicles, a distance constraint is established between AVs, defined by a constant, $\Delta$, such that $p_{k}^{\mathbf{n_a}-1}-p_{k}^{\mathbf{n_a}} > \Delta$. Given the uncertain nature of the HV behavior model, a probabilistic constraint, or chance constraint, is set to ensure a safe distance between the trailing AV and the HV. This constraint is expressed as:
\begin{equation} 
    \mathrm{Pr}\left(p_{k}^{N_a}-(p_{k}^{H} + \Delta) > \Delta_\text{ext} \right) \geq p_{\text{def}} \, , \tag{13} \label{eqn:chance_constraint} 
\end{equation}
Here, $\Delta$ denotes the predefined safe distance between AVs, and $\Delta_\text{ext} \geq 0$ is an extra distance established to account for the HV's stochastic behavior. The required satisfaction probability is denoted as $p_{\text{def}}$. To rephrase the distance constraint $\mathcal{X}$, we can express it as a single half-space constraint $\mathcal{X}^{hs} := \bigl\{x \vert h^{\top}x \leq b \bigl\}$, $ h \in \mathbb{R}^n$, where $ h \in \mathbb{R}^n$, and $b \in \mathbb{R}$. As derived by Hewing et al. in \citep{hewing2019}, a method to tighten the constraint on the state mean is expressed as:
\begin{equation} 
    \mathcal{X}^{h s}\left(\Sigma_{i}^{x}\right):=\left\{x \mid h^{\top} x \leq b-\phi^{-1}\left(p_{\text{def}}\right) \sqrt{h^{\top} \Sigma_{i}^{x} h}\right\} \, , \tag{14} \label{eqn:18} 
\end{equation}
In this equation, $h^{\top} = \begin{bmatrix} -1 & 1\end{bmatrix}$, $x := \begin{bmatrix} p_{k}^{N_a} & p_{k}^{H}+\Delta \end{bmatrix}^{\top}$, and $b = -\Delta_\text{ext}$. Here, $\phi^{-1}$ is the inverse of the cumulative distribution function (CDF). In our scenario:
\begin{equation} 
    \Sigma_{k}^{x} := \begin{bmatrix} \Sigma_{k}^{p^{N_a}} \\ \Sigma_{k}^{p^{H}}+\Delta \end{bmatrix} = \begin{bmatrix} 0 & 0 \\ 0 & \Sigma_{k}^{p^{H}}\end{bmatrix} \, . \tag{15} \label{eqn:19} 
\end{equation}
Here, there is no covariance between the position of the HV and the AV leading it, indicated by $\Sigma_{k}^{p^{N_a}} = 0$. The position variance of the HV, denoted as $\Sigma_{k}^{p^{H}}$, is computed using \eqref{eqn:variance_prop_b}. To establish a more ``tightened'' constraint on the position state, we substitute \eqref{eqn:18} into \eqref{eqn:19} to obtain:
\begin{equation} 
    p_{k}^{N_a}-p_{k}^{H} \geq \Delta + \Delta_\text{ext} + \phi^{-1}\left(p_{\text{def}}\right) \sqrt{ \Sigma_{k}^{p^{H}}} \, . \tag{16} \label{eqn:safe_dis_hv} 
\end{equation}

The safe distance probability constraint outlined in \eqref{eqn:chance_constraint} is simplified as a deterministic equation described in \eqref{eqn:safe_dis_hv}. The additional term, $\Delta_\text{ext}$, can be set to zero when a high level of assurance in constraint satisfaction is desired, leading to further simplification. The AV--HV safe distance is adaptively adjusted, leveraging the HV's uncertainty estimates from the Gaussian Process (GP) model, as indicated in \eqref{eqn:variance_prop_b}. This adjustment ensures the safe distance requirement exceeds $\Delta$ under all conditions.

\subsection{GP learning-based MPC}
\label{sec:GP-MPC}
We designed an MPC strategy that incorporates the proposed modeling method for the human-driven vehicle (HV) in a mixed-vehicle platoon scenario. The platoon comprises $N_a$ autonomous vehicles (AVs) and one HV, as shown in Figure \ref{figure:1}. The strategy is represented as follows:
\begin{ceqn}
    \begin{align} 
       \underset{\mathbb{V}}{\text{min}} \sum_{\mathbf{n_a}=1}^{N_a} & \sum_{i=k}^{k+N-1} \Big\| \mathrm{acc}_{{i}|k}^{\mathbf{n_a}} \Big\|^2_R + \sum_{i=k}^{k+N} \Big\| v _{{i+1}|k}^1 - v_{{i+1}|k}^\text{ref} \Big\|^2_{Q_1} \nonumber \\
       & \qquad \quad + \sum_{\mathbf{n_a}=2}^{N_a}\sum_{i=k}^{k+N} \Big\| v_{{i+1}|k}^{\mathbf{n_a}} - v_{{i+1}|k}^{\mathbf{n_a}-1} \Big\|^2_{Q_2} \tag{17a} \label{eqn:mpc_a} \\
       \text{with} \ \mathbb{V} = & \left\{v_{{i}|k}^1, v_{{i}|k}^{\mathbf{n_a}}, v_{{i}|k}^{H}, p_{{i}|k}^{\mathbf{n_a}}, \mu_{{i}|k}^{p^{H}}, \Sigma_{{i}|k}^{p^{H}}, \mathrm{acc}_{{i}|k}^{\mathbf{n_a}} \right\} \, \nonumber \\
        \text {subject to} \nonumber \\
        v_{{i+1}|k}^{\mathbf{n_a}} &= v_{{i}|k}^{\mathbf{n_a}} + T \, \mathrm{acc}_{{i}|k}^{\mathbf{n_a}} , \  p_{{i+1}|k}^{\mathbf{n_a}} = p_{{i}|k}^{\mathbf{n_a}} + T \, v_{{i}|k}^{\mathbf{n_a}} \, ,\tag{17b} \label{eqn:mpc_b} \\ 
        v^{H}_{{i}|k} &= {f}\left(v_{i-1:i-4 | k}^{H}, {v}_{i-1:i-4 | k}^{N_a} \right) \, , \tag{17c} \label{eqn:mpc_c}\\
        \mu_{{i+1}|k}^{p^{H}} &= \mu_{{i}|k}^{p^{H}} + T \, v_{{i}|k}^{H} + T \, \mu^d (v_{{i-1}|{k}}^{H}, {v}_{{i-1}|{k}}^{N_a}) \, , \tag{17d} \label{eqn:mpc_d}\\
        \Sigma_{{i+1}|k}^{p^{H}} &= \Sigma_{{i}|k}^{p^{H}} + T^2 \Sigma^d (v_{{i-1}|{k}}^{H}, {v}_{{i-1}|{k}}^{N_a}) \, , \tag{17e} \label{eqn:mpc_e}\\
        p_{{i}|k}^{\mathbf{n_a}-1} & - p_{{i}|k}^{\mathbf{n_a}} \geq \Delta \, , \tag{17f} \label{eqn:mpc_f}\\
        p_{{i}|k}^{N_a} & - \mu_{{i}|k}^{p^{H}} \geq \Delta + \phi^{-1}\left(p_{\text{def}}\right) \sqrt{ \Sigma_{{i}|k}^{p^{H}}} \, , \tag{17g} \label{eqn:mpc_g}\\
        v_{\text{min}} & \leq v_{{i}|k}^{\mathbf{n_a}} \leq v_{\text{max}} \, , \ \mathrm{acc}_{\text{min}} \leq \mathrm{acc}_{{i}|k}^{\mathbf{n_a}} \leq \mathrm{acc}_{\text{max}} \, . \tag{17h} \label{eqn:mpc_h}
    \end{align}
\end{ceqn}
%
% The current time step is $k$, and the system (whether hardware or simulation) commences at $k=0$. In the MPC prediction horizon, the prediction horizon time step originates at $i=1$, with variable values at $i=0$ initialized using measurements.

Due to the unpredictable nature of human drivers, the cost function \eqref{eqn:mpc_a} does not incorporate HV velocities. Instead, it centers on the discrepancies between the reference velocity and the lead AV's velocity, velocity differences among adjacent AVs, and control inputs, with each component weighted by positive constants $Q_1$, $Q_2$, and $R$. The HV velocities, however, are included in the GP+ARX HV model, which introduces the compensated HV velocities in \eqref{eqn:mean_prop} and propagates HV variances as depicted in \eqref{eqn:variance_prop_b}. Additionally, an adaptive AV--HV safe distance is integrated, utilizing the HV variance propagation from \eqref{eqn:safe_dis_hv}. These elements are explicitly considered as equality and inequality constraints in the formulation of the MPC policy.

More specifically, the equality constraint \eqref{eqn:mpc_b}$-$\eqref{eqn:mpc_e} includes the AV model as specified in \eqref{eqn:av_eqn_a} and \eqref{eqn:av_eqn_b}, along with the ARX model of the HV, represented by \eqref{eqn:arx}, as stated in \eqref{eqn:mpc_c}. The HV mean and variance position propagation equations, which were derived in \eqref{eqn:mean_prop} and \eqref{eqn:variance_prop_b} respectively, are represented by \eqref{eqn:mpc_d} and \eqref{eqn:mpc_e}. In these equations, $\mu^d(\cdot)$ and $\Sigma^d(\cdot)$ correspond to the mean and variance predictions of the GP model that is defined in \eqref{eqn:gp_eqn}. The inequality constraints include the velocity and acceleration conditions \eqref{eqn:mpc_h}, as well as the safe distance requirements \eqref{eqn:mpc_f} and \eqref{eqn:mpc_g}.

\subsection{Dynamic sparse GP prediction in MPC}
\label{sec:dynamic_sparseGP_mpc}
In Section \ref{sec:sparse_hv}, we demonstrated the effectiveness of the sparse GP+ARX model in reducing computation time while improving modeling accuracy for HV velocity predictions compared to the standard GP+ARX and ARX models. However, integrating GP models into the MPC framework to achieve real-time calculations remains a challenge. MPC itself is computationally expensive as it involves solving an optimal control problem at each time step while satisfying constraints specified in equations \eqref{eqn:mpc_b}--\eqref{eqn:mpc_h}. To address this challenge, we implemented a dynamic sparse approximation for the GP-MPC method proposed in \citep{hewing2019} to achieve further speed-up.

Given the receding horizon feature of MPC, the predictive trajectory at the current time step is similar to the trajectory calculated at the previous time step. Leveraging this observation, we applied the sparse GP to perform calculations on the trajectory obtained at the previous time step. We then used the results to propagate the mean and variance at the current time step by utilizing equations \eqref{eqn:mpc_d} and \eqref{eqn:mpc_e}. This approach reduces the computational burden by using a single sparse GP prediction for the entire prediction horizon instead of individual predictions for each time step within the horizon. Consequently, it enables the GP-MPC scheme to be computationally feasible for real-time applications with fast sampling rates. 

\section{Simulations}
\label{sec:simulations}

This section aims to elucidate the enhanced control performance gained through employing the proposed HV model. We conducted a simulated case study where the control performance of a mixed vehicle platoon implementing the sparse GP-based MPC strategy, as delineated in Sec. \ref{sec:GP-MPC}, and a standard MPC were juxtaposed quantitatively. The simulations were all conducted using MATLAB R2022a on an Ubuntu 20.04 laptop. The source code is available at the following repository: \url{https://github.com/CL2-UWaterloo/GP-MPC-of-Platooning}. It's noteworthy that the AV platoon contained two HVs in all the presented case studies. However, the algorithm and codes are engineered to be applicable and scalable to platoons comprising more than two AVs.

To maintain consistency in the initial conditions for all simulations, the initial velocities of each vehicle in the platoon were kept at zero. Both the standard MPC and the sparse GP-based MPC used a sample time of T = 0.25 s and a predictive horizon of N = 6. The cost function weights for both MPC policies were adjusted to $Q_1 = Q_2 = 5$ and $R = 20$. We defined the maximum acceleration as $a_{\text{max}} =$ 5\,m$^2/$s and the minimum acceleration as $a_{\text{min}} =$ -5 m$^2/$s. Additionally, we set the maximum velocity as $v_{\text{max}} =$ 35\,m$/$s and the minimum velocity as $v_{\text{min}} =$ -35\,m$/$s. The minimum safe distance was set to $\Delta =$ 20\,m in equations \eqref{eqn:mpc_f} and \eqref{eqn:mpc_g}. For the initial positions in our simulations, we adopted the following configuration: the first leader AV was positioned at $p = 0$, the second AV was located $1.2\Delta =$ 24\,m behind the first leader AV, and the HV was situated $1.2 \Delta =$ 24\,m behind the second AV. This initial arrangement allowed for the desired spacing between the vehicles in the platoon. The chance constraint satisfaction probability in \eqref{eqn:chance_constraint} was fixed at $p_{\text{def}}=0.95$.

To set a benchmark for comparison, we employed a standard MPC, hereafter referred to as the nominal MPC. This controller utilizes the HV's ARX model as outlined in equations \eqref{eqn:mpc_c} and \eqref{eqn:mpc_d} in its prediction loop, and excludes the third component of the GP model. The distance constraint between the HV and the second AV, defined as $\Delta$, does not account for the adaptive component in equation \eqref{eqn:mpc_g}. Moreover, the nominal MPC does not consider position variance propagation as per equation \eqref{eqn:mpc_e}. We employed the ARX+GP model, as derived in equations \eqref{eqn:sys_model_a} and \eqref{eqn:sys_model_b}, to simulate the HV model in both the nominal MPC and the sparse GP-based MPC.

\begin{figure}
    \centering
    \vspace{0.2cm}
    \subfloat{{\includegraphics[trim=0cm 0cm 0cm 0cm, width=0.94\columnwidth]{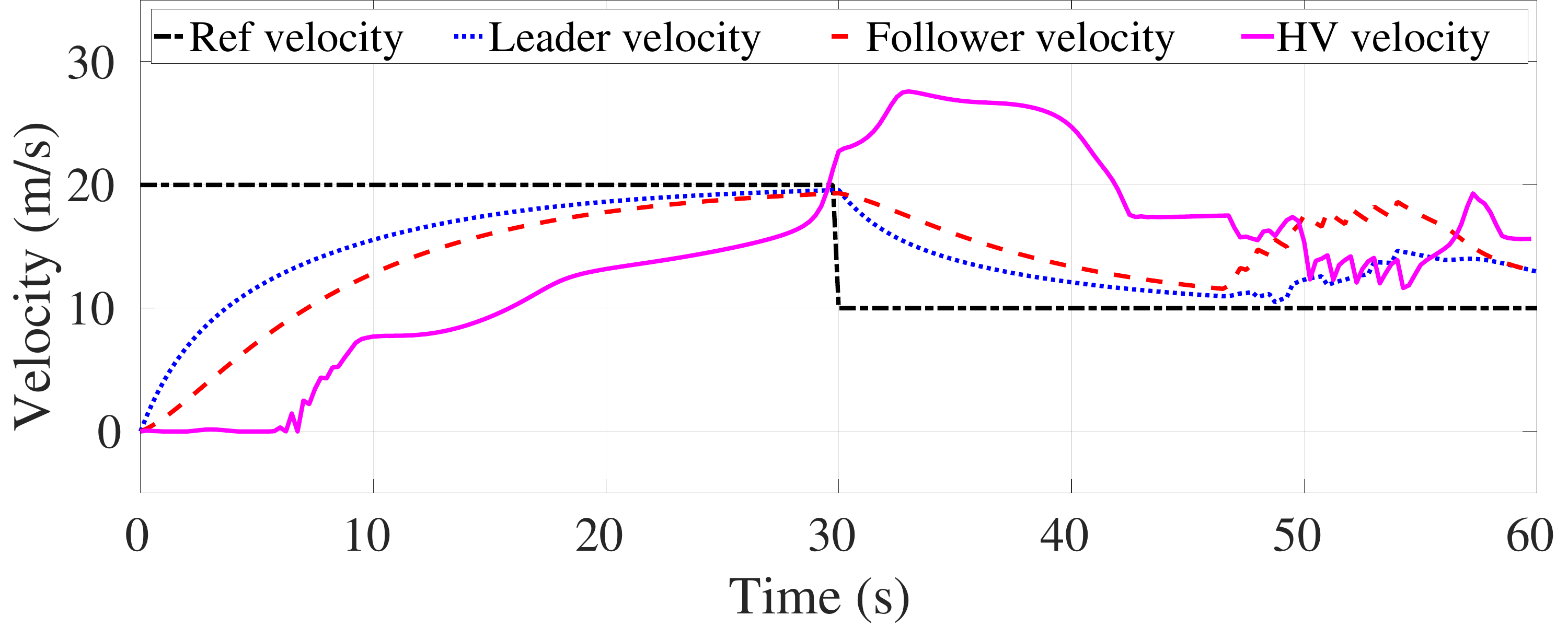} }}
    \qquad \qquad 
    \subfloat{{\includegraphics[trim=0cm 0cm 0cm 0.0cm, width=0.94\columnwidth]{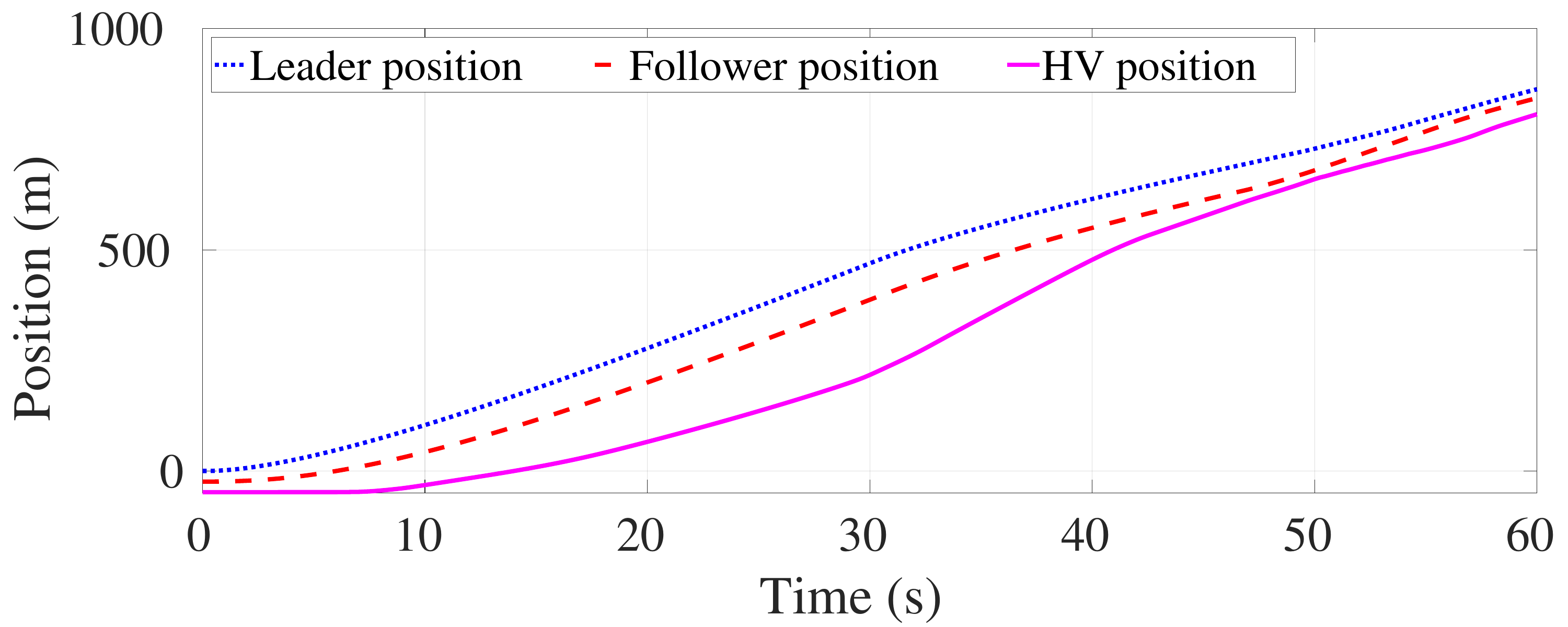} }}
    \qquad \qquad 
    \vspace{0.05cm}
    \subfloat{{\includegraphics[trim=0.0cm 0cm 0.0cm 0.0cm, width=0.94\columnwidth]{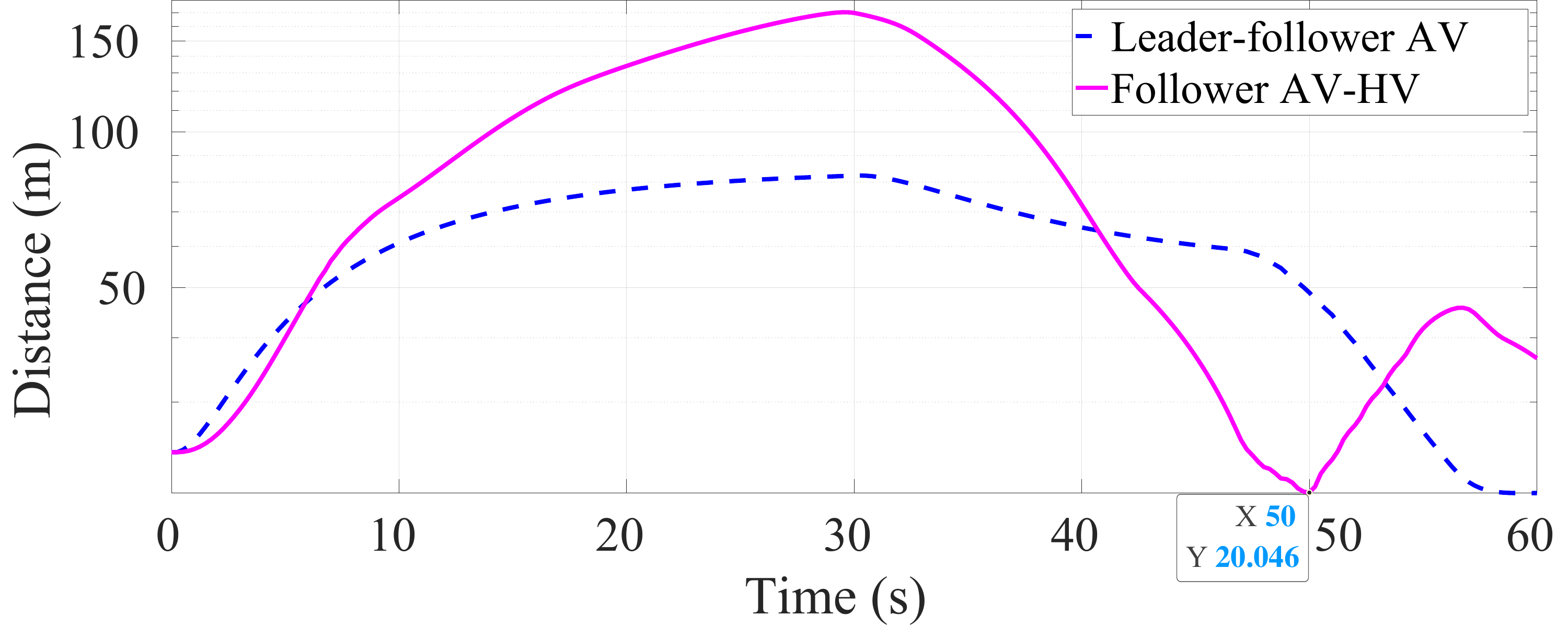} }}
    % \qquad \qquad 
    % \subfloat{{\includegraphics[trim=0.0cm 0cm 0.0cm 0.0cm, width=0.96\columnwidth]{figures/20-10_s_control_nominal_60s.pdf} }}
    \caption{The simulation results for an emergency braking scenario using the nominal MPC. The figures, arranged in descending order, correspond to the velocity tracking, vehicle positioning, and relative distance between the vehicles, respectively. 
    }
\label{figure:nominal_simulation_braking}
\end{figure}

\begin{figure}
    \centering
    \vspace{0.2cm}
    \subfloat{{\includegraphics[trim=0cm 0cm 0cm 0cm, width=0.94\columnwidth]{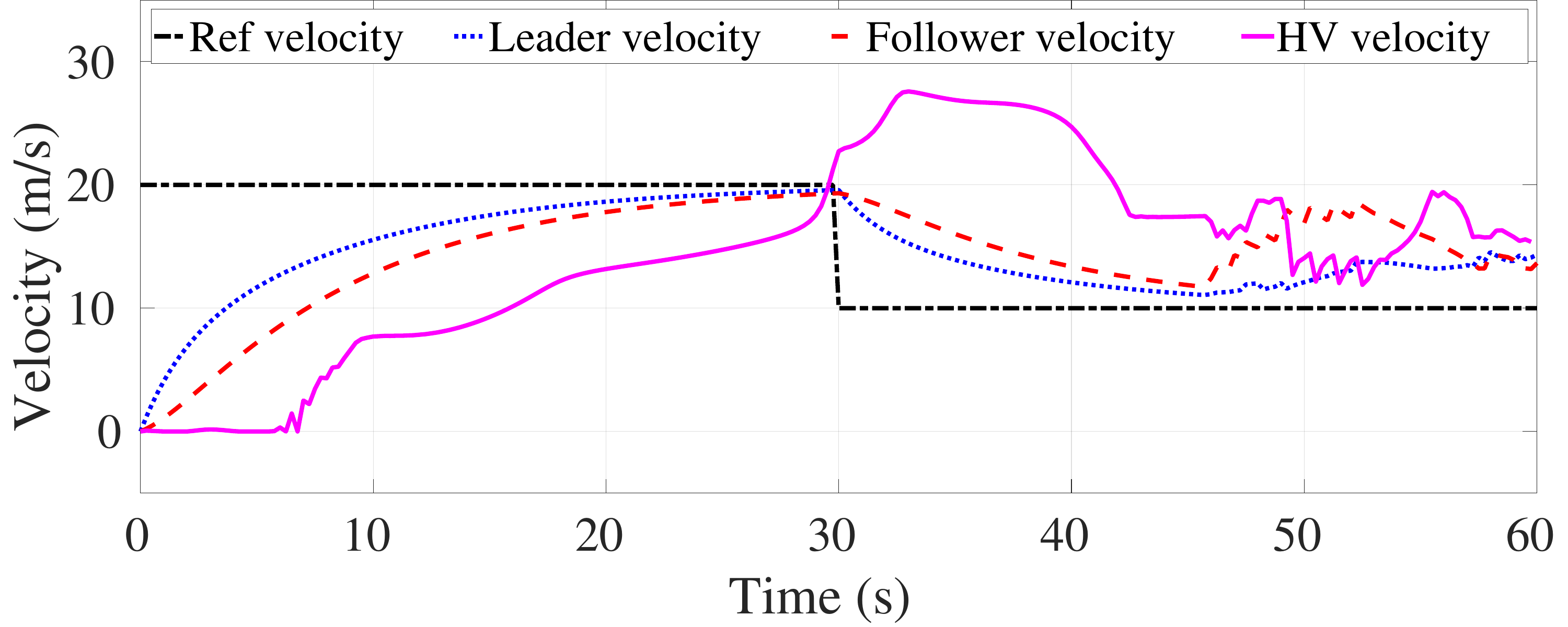} }}
    \qquad \qquad 
    \subfloat{{\includegraphics[trim=0cm 0cm 0cm 0.0cm, width=0.94\columnwidth]{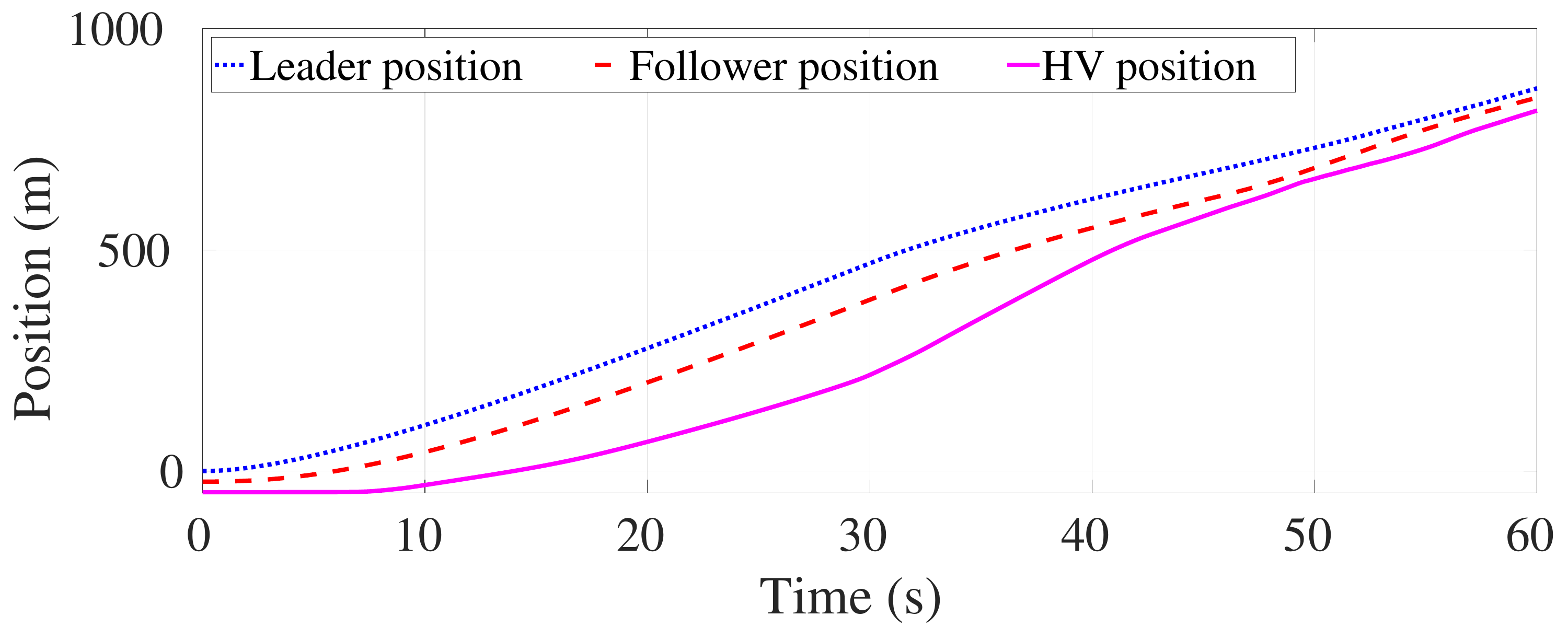} }}
    \qquad \qquad 
    \vspace{0.05cm}
    \subfloat{{\includegraphics[trim=0cm 0cm 0cm 0.0cm, width=0.94\columnwidth]{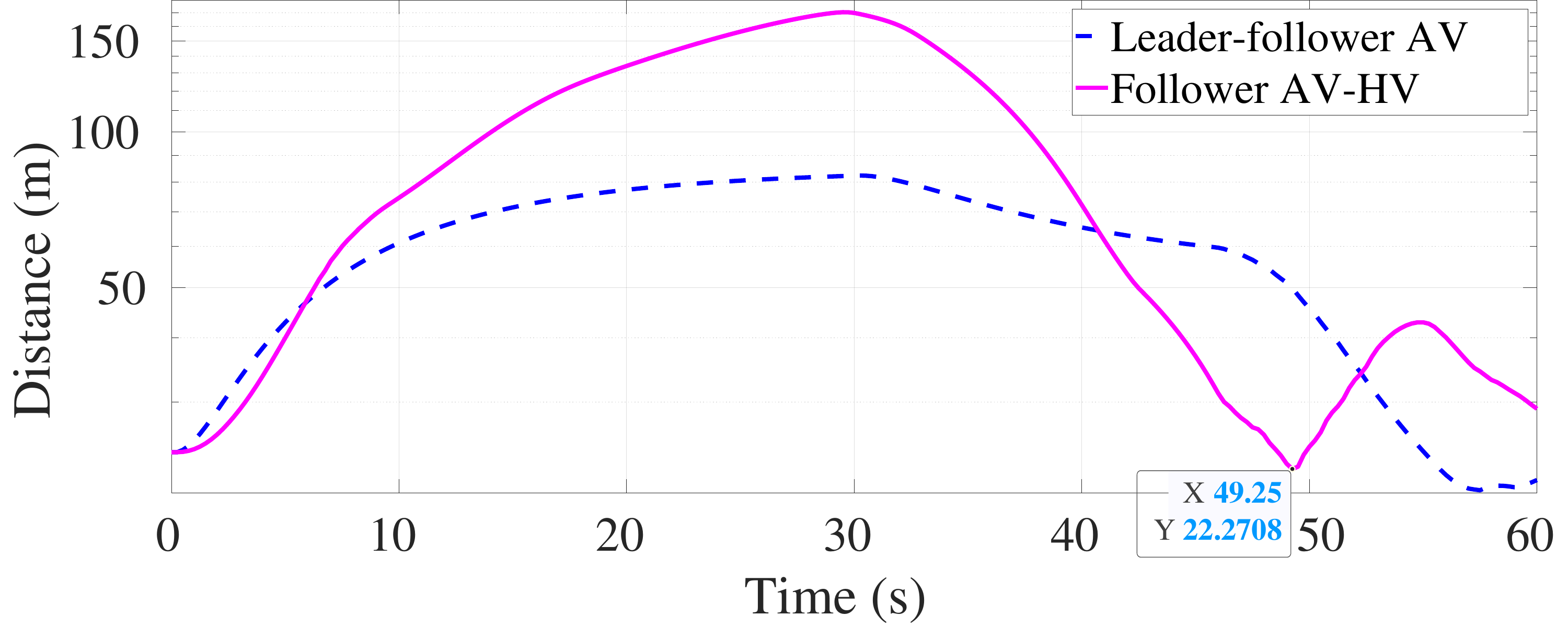} }}
    % \qquad \qquad 
    % \subfloat{{\includegraphics[trim=0.0cm 0cm 0.0cm 0.0cm, width=0.96\columnwidth]{figures/20-10_s_control_60s.pdf} }}
    \caption{The results of simulating an emergency braking scenario using the proposed GP-MPC are presented across several plots. These plots, sequentially representing velocity tracking, vehicle positioning, and relative distance between vehicles, depict the dynamic interactions within the platoon. In comparison with the baseline MPC, as illustrated in Fig. \ref{figure:nominal_simulation_braking}, the GP-MPC affords a two-meter increase in the minimum distance between the HV and the trailing AV. Moreover, as depicted in Tab. \ref{tab:simulation_metrics}, all vehicles cover larger distances under the GP-MPC model. Thus, the GP-MPC approach enhances safety without enforcing more stringent operational constraints than the nominal MPC, given that all vehicles travel a larger distance.}
    \label{figure:gp_simulation_braking}
\end{figure}

We performed a case study simulating an emergency braking scenario, using both the nominal MPC and the proposed GP-MPC policy. The scenario involved a leading AV with a reference velocity of $v^\text{ref}=$ 20\,m$/$s, which was suddenly reduced to $v^\text{ref}=$ 10\,m$/$s at $t=$ 30\,s. This particular emergency braking scenario was selected to thoroughly evaluate the robustness and efficacy of our HV modeling method and control policy under strenuous conditions. Additionally, the primary driving factor behind this study was to enhance safety in HV--AV platooning by minimizing the chances of rear-end accidents involving HVs and AVs. Subsequently, we generated plots depicting the velocity tracking, position of each vehicle, and relative distance between the vehicles, as presented in Fig. \ref{figure:nominal_simulation_braking} for the nominal MPC and in Fig. \ref{figure:gp_simulation_braking} for the GP-MPC. The top plots of the velocity tracking exemplify the cooperative maneuvers of the leader and follower AVs. Both AVs accelerated at approximately $t=$ 46\,s to ensure adherence to the safe distance constraint.


In Tab. \ref{tab:simulation_metrics}, we provided a quantitative representation of the performance enhancements offered by the proposed GP-MPC. The GP-MPC attained a minimum HV--AV distance of 22.27\,m, which is two meters more than the nominal MPC. This enhancement is attributed to the incorporation of an additional component in the safe distance constraint that accounts for the GP uncertainty assessments as defined in \eqref{eqn:safe_dis_hv}. Additional distances were appended to the distance constraint in \eqref{eqn:mpc_g} at every time step in the GP-MPC to better ensure safety. All vehicle positions in the mixed platoon of the GP-MPC were ahead of the nominal MPC, suggesting that the GP-MPC also achieved a higher speed for each vehicle in the mixed platoon.

\begin{table}
    \caption{Simulation performance metrics: The position of all vehicles and the minimum relative distance of HV--AV.
    % Compared to the nominal MPC, bigger position values for the GP-MPC indicate a higher overall speed of the mixed vehicle platoon, and a larger Min HV--AV distance provides an extra safety guarantee.
    }
    \label{tab:simulation_metrics}
    \begin{center}
        \begin{tabular}{|c|c|c|c|c|}
            \hline Controller & AV 1 & AV 2 & HV & Min HV--AV  \\
            \hline Nominal MPC & 862.90\,m & 842.89\,m & 806.42\,m & 20.05\,m\\
            \hline GP-MPC & \textbf{864.92}\,m & \textbf{843.72}\,m &  \textbf{814.60}\,m & \textbf{22.27}\,m\\
            \hline
        \end{tabular}
    \end{center}
\end{table}

We also compared the computation time of the GP-MPC and the nominal MPC at each time step in Tab. \ref{tab:simulation_time}. The average computation time (Ave Time) of the GP-MPC is only 5\% more than that of the nominal MPC. The maximum time (Max Time) of the GP-MPC shows that it can operate in real-time at 4 HZ, and the small standard deviation (Time Std) of the GP-MPC indicates a more consistent computation time compared to the nominal MPC. Compared to our previous work \citep{wang2024improving} that did not employ sparse GP modeling for HVs and the dynamic GP prediction technique in MPC, the average computation time per time step for the GP-MPC has been reduced by approximately 100 times.

\begin{table}
    \vspace{0.2cm}
    \caption{Simulation computation time and accumulative control cost: The averaged time at each time step of the GP-MPC is only 5\% more than the nominal MPC. The accumulative control cost of the GP-MPC is 0.9\% more than the nominal MPC.
    % The smaller standard deviation (Time Std) value of the GP-MPC indicates a more consistent computation time than the nominal MPC. 
    }
    \label{tab:simulation_time}
    \begin{center}
        \begin{tabular}{|c|c|c|c|c|}
            \hline Controller & Ave Time & Max Time & Time Std & Control Cost \\
            \hline Nominal MPC & \textbf{0.2062}\,s & \textbf{0.2347}\,s & 0.065 & \textbf{51329} \\
            \hline GP-MPC & 0.2186\,s & 0.2391\,s & \textbf{0.006} & 51805 \\
            \hline
        \end{tabular}
    \end{center}
\end{table}

To offer a more comprehensive comparison between the proposed GP-MPC and the nominal MPC, we introduced an accumulative control cost metric. This metric was derived using a modified cost function, which is based on \eqref{eqn:mpc_a}, and was specifically tailored for the vehicle platoon simulations:
\begin{equation}
    \sum_{k=1}^{T} \left(\| v_{k}^1 - v_{k}^\text{ref} \|^2_{Q_1} + \| v_{k}^{2} - v_{k}^{1} \|^2_{Q_2} \right) + \sum_{k=1}^{T-1} \left( \| a_{k}^{1} \|^2_R + \| a_{k}^{2} \|^2_R \right) \, . \tag{18} \label{eqn:mpc_cost_total}
\end{equation} 
The results of the accumulative control cost can be found in Tab. \ref{tab:simulation_time}. From the results, it's evident that while the proposed GP-MPC offers superior safety guarantees (evidenced by the greater minimum distance between HV--AV) and demonstrates heightened travel efficiency (demonstrated by the extended traveled distance achieved by all vehicles), it does demand 0.92\% more control effort than the nominal MPC. One potential avenue for improvement is the optimization of the control gains of the MPC policy, although this wasn't the primary focus of our current study. Notably, when comparing the traveled distance of the HV using both the nominal MPC and the proposed GP-MPC, the latter yields a 1\% improvement in distance. Consequently, while the proposed GP-MPC might necessitate greater control effort and potentially increased power consumption, it offsets this by offering a more significant percentage improvement in travel efficiency.

\section{Conclusion}
\label{sec:conclusion}
% Follow \url{https://people.eecs.berkeley.edu/~anca/papers/IROS16_active.pdf}

This paper presents an innovative learning-based modeling approach for human-driven vehicles, combining a first-principles nominal model with a Gaussian process learning-based component. The proposed model improves modeling accuracy and estimates uncertainty in the HV model, enabling enhanced control of mixed vehicle platoons. Based on this model, a model predictive control strategy is developed for longitudinal car-following scenarios in mixed platoons. The developed MPC policy is evaluated through simulation case studies, comparing it to a baseline MPC approach. The results demonstrate that our approach provides superior safety guarantees and facilitates a more efficient motion behavior for each vehicle in the mixed platoon. By employing a dynamic sparse GP technique within each MPC prediction loop, our MPC achieves computational efficiency, requiring only 5\% more computation time compared to the baseline nominal MPC. This represents a significant reduction in computation time compared to our previous work \citep{wang2024improving}, with an approximate 100-fold improvement.

In this work, we emphasize the potential and practicality of the GP-based MPC by utilizing the proposed HV model to bolster safety in mixed-vehicle platoons. A comprehensive formal stability analysis, particularly local stability and string stability \citep{pirani2022}, is not presented due to the inherent complexities of the approach. GP-based MPC, although increasingly prominent, presents unresolved challenges in stability assertions \citep{berberich2020}. Currently, adaptive MPC frameworks, inclusive of GP-based methodologies, lack guarantees for recursive feasibility and stability, especially when dealing with nonlinear system dynamics under probabilistic constraints \citep{bujarbaruah2020}. The nonlinearity of GP models, which are adept at capturing underlying system dynamics, complicates the application of traditional linearized stability analyses. Additionally, when considering interconnected systems or platoons, the behavior dependencies introduced by GP models further challenge the direct application of string stability principles. Such interdependencies, coupled with the variability of individual GP models in a platoon scenario, render the conventional string stability methods less insightful \citep{zhang2022}. Given these complexities, our methodology prioritizes empirical validation through simulations, which effectively demonstrates the system's adherence to constraints and its capability to maintain safe operational behaviors \citep{hewing2020, haninger2022}. 

Our work has some limitations that require further research. Firstly, it is limited to longitudinal car-following scenarios, specifically a human-driven vehicle following an autonomous vehicle platoon. Expanding the approach to more complex traffic scenarios, such as human-driven vehicles within the platoon or involving merging and lane-changing maneuvers, would be valuable. Secondly, the restricted diversity of the collected data used to train the GP models for human-driven vehicles suggests the potential for improving model accuracy by including more diverse drivers and driving scenarios. Therefore, collecting data and assessing performance on urban networks, which feature frequent speed variations beyond the examined speed profile, would be a compelling future direction. Additionally, conducting further research on the stability analysis of GP-based MPC would advance the theoretical understanding and enable broader application of GP-based MPC in real-world scenarios.

\section{Acknowledgements}

We would like to express our gratitude for the financial support provided by the Natural Sciences and Engineering Research Council (NSERC) of Canada through the Discovery Grant program and Magna International. 

% %% If you have bibdatabase file and want bibtex to generate the
% %% bibitems, please use
% %%
% \clearpage
% \bibliographystyle{elsarticle-num} 
\bibliographystyle{elsarticle-harv}
\bibliography{bibliography}

% %% else use the following coding to input the bibitems directly in the
% %% TeX file.

% % \begin{thebibliography}{00}

% % %% \bibitem{label}
% % %% Text of bibliographic item

% % \bibitem{}

% % \end{thebibliography}

\end{document}